%% file: arxiv.tex
\documentclass[letterpaper]{article} 
\usepackage{aaai2026}  
\usepackage{times}  
\usepackage{helvet}  
\usepackage{courier}  
\usepackage[hyphens]{url}  
\usepackage{graphicx} 
\usepackage{natbib}  
\usepackage{caption} 
\usepackage{algorithm}
\usepackage{algorithmic}

\usepackage{multirow}
\usepackage{booktabs}
\usepackage{pifont}
\newcommand{\cmark}{\checkmark}
\newcommand{\xmark}{\ding{55}}
\usepackage{amsfonts}
\usepackage{amsmath}

\usepackage{newfloat}
\usepackage{listings}
\DeclareCaptionStyle{ruled}{labelfont=normalfont,labelsep=colon,strut=off} 
\floatstyle{ruled}
\newfloat{listing}{tb}{lst}{}
\floatname{listing}{Listing}
\newcommand{\name}{ManipDreamer3D }
\title{
ManipDreamer3D: Synthesizing Plausible Robotic Manipulation Video with Occupancy-aware 3D Trajectory
}

\author{
    Ying Li\textsuperscript{\rm 1,2,3,4},
    Xiaobao Wei\textsuperscript{\rm 1,3},
    Xiaowei Chi\textsuperscript{\rm 2}\textsuperscript{\dag},
    Yuming Li\textsuperscript{\rm 1,4},
    Zhongyu Zhao\textsuperscript{\rm 1,4},\\
    Hao Wang\textsuperscript{\rm 1},
    Ningning Ma\textsuperscript{\rm 3},
    Ming Lu\textsuperscript{\rm 1}\textsuperscript{\ddag},
    Sirui Han\textsuperscript{\rm 2}\textsuperscript{\ddag},
    Shanghang Zhang\textsuperscript{\rm 1}\textsuperscript{\ddag},
}
\affiliations{
    \textsuperscript{\rm 1}State Key Laboratory of Multimedia Information Processing, School of Computer Science, Peking University\\
    \textsuperscript{\rm 2}Hong Kong University of Science and Technology
    \textsuperscript{\rm 3}Autonomous Driving Development, NIO\\
    \textsuperscript{\rm 4}School of Software and Microelectronics, Peking University\\
    \textsuperscript{\dag}Project Leader \quad
    \textsuperscript{\ddag}Corresponding Authors\\
}

\usepackage{bibentry}

\begin{document}

\maketitle

\begin{abstract}
    Data scarcity continues to be a critical bottleneck in the field of robotic manipulation, limiting the ability to train robust and generalizable models. While diffusion models provide a promising approach to synthesizing realistic robotic manipulation videos, their effectiveness hinges on the availability of precise and reasonable control instructions. Current methods primarily rely on 2D trajectories as instruction prompts, which inherently face issues with 3D spatial ambiguity. 
    In this work, we present a novel framework named \name for generating plausible 3D-aware robotic manipulation videos from the input image and the text instruction. Our method combines 3D trajectory planning with a reconstructed 3D occupancy map created from a third-person perspective, along with a novel trajectory-to-video diffusion model. 
    Specifically, \name first reconstructs the 3D occupancy representation from the input image and then computes an optimized 3D end-effector trajectory, minimizing path length, avoiding collisions and retiming. Next, we employ a latent editing technique to create video sequences from the initial image latent, text instruction and the optimized 3D trajectory. This process conditions our specially trained trajectory-to-video diffusion model to produce robotic pick-and-place videos. Our method significantly reduces human intervention requirements by autonomously planing plausible 3D trajectories. Experimental results demonstrate its superior visual quality and precision.
\end{abstract}

\input{sections/1_introduction}
\input{sections/2_related_works}
\input{sections/3_method}
\input{sections/4_experiments}
\input{sections/5_conclusions}

\newpage
\section{Acknowledgments}
This work was supported by the National Natural Science Foundation of China (62476011).

\bibliography{aaai2026}

\end{document}

%% file: sections/1_introduction.tex
\section{Introduction}
\label{sec:paper-introduction}

Collecting real-world robot manipulation demonstrations is often time-consuming, labor-intensive, and constrained by hardware limitations~\cite{park2024dexhub, an2025dexterous}. These challenges hinder the scalability of robotic policy learning, where large and diverse datasets are critical for achieving robust generalization~\cite{schmidt2018adversarially, xu2012robustness}. Generating realistic demonstrations with safe and short trajectories becomes particularly important~\cite{ding2020learning, hanselmann2022king}, as it can reduce the dependency on extensive physical data collection while providing high-quality supervision for an effective robot policy model. 

Recent works have sought to expand real-world robot manipulation datasets by replaying recorded trajectories with diverse visual augmentations, such as altering objects, robot embodiments, textures, backgrounds, distractors, lighting conditions, and camera viewpoints~\cite{mandi2022cacti, Chen-RSS-23, fang2025rebot, yang2025novel}. 
Beyond augmentation-based replay, Re$^{3}$Sim~\cite{han2025re} reconstructs realistic scenes in simulation to synthesize plausible manipulation demonstrations, while ORV~\cite{yang2025orv} uses 4D semantic occupancy as an intermediate representation to generate realistic videos from either real-world or simulated data. 
Modeling robotic actions serves as another vital method for enriching realistic manipulation datasets. This\&That~\cite{wang2024language} conditions generation on object and target gestures, whereas RoboMaster~\cite{fu2025learning} jointly models both robot and object trajectories to guide the generation process. 
\begin{figure*}[ht]

\centering

\includegraphics[width=0.9\textwidth]{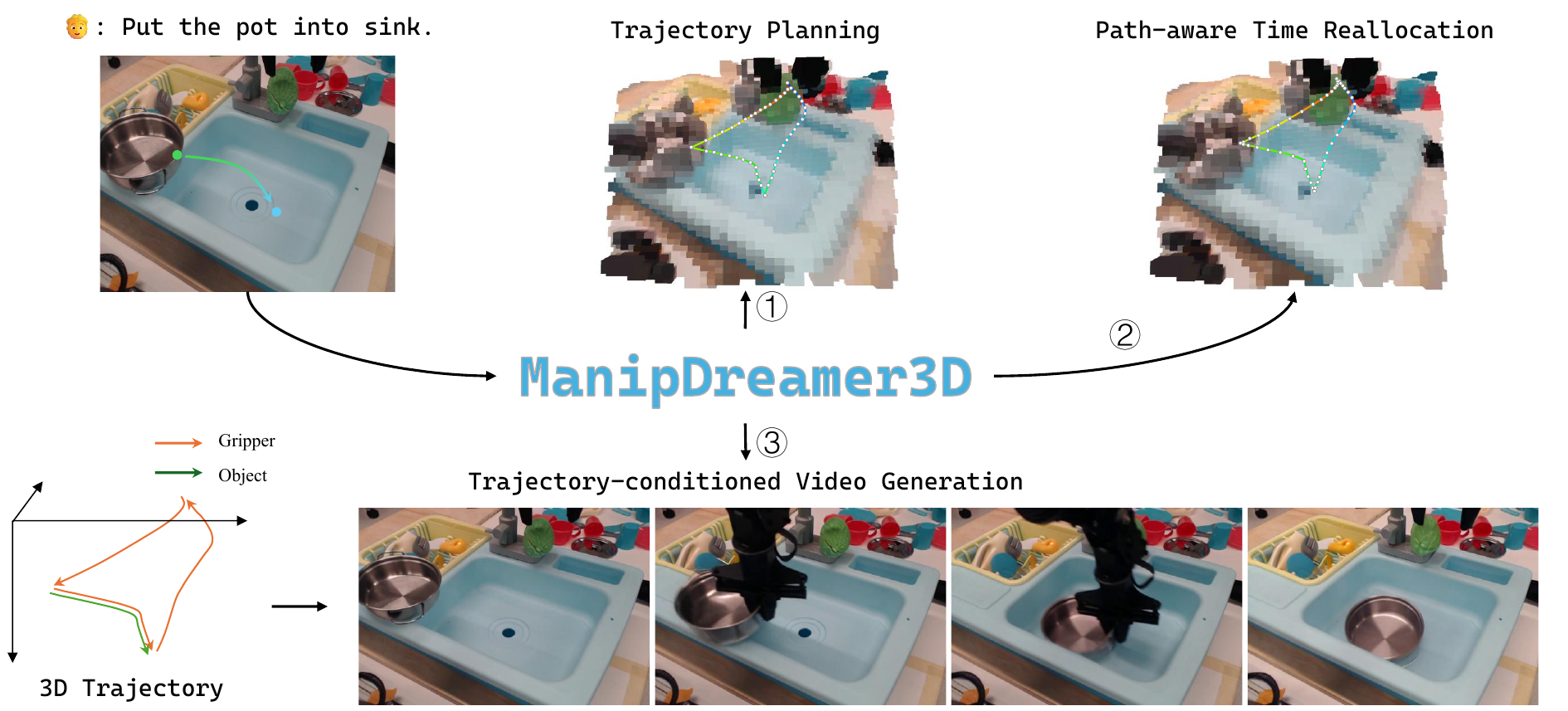}

\caption{\textbf{Overview of our proposed method \name.} Given a user-specified third-view image, an instruction, and gestures, \name first constructs an occupancy grid and initializes and optimizes sub-trajectories within the grid space. The time intervals are re-allocated based on the sub-trajectory path lengths and predefined velocity profiles. Finally, \name synthesizes the output video conditioned on the trajectories of both the robot end-effector and the object.}

\label{fig:overview}

\end{figure*}

While these approaches have generated promising robotic videos, several limitations remain.
First, existing methods often overlook that robotic actions are planned in a 3D space. As a result, the generated videos may contain trajectories that violate physical constraints, lack collision avoidance, or execution efficiency. These methods still rely heavily on manual object selection~\cite{yang2025orv, wang2024language, fu2025learning}. 
Second, the generated scenes fail to maintain geometric and physical consistency with real-world environments~\cite{wang2024language, zhou2024robodreamer, li2025manipdreamer}. Even when the generated 2D videos exhibit promising perceptual quality, inaccuracies in object size, placement, or contact state can lead to unrealistic interactions. These issues significantly limit the applicability of such videos for training generalizable robotic policies. 

To overcome these limitations, we propose \name, a novel method to automatically generate both realistic and physically-plausible robotic manipulation videos given a third-view observation image along with an instruction. 
The key idea is to first reconstruct a 3D occupancy representation of the scene and plan a physically valid, short manipulation trajectory with reasonable speed from the robot end-effector to the manipulated object, and then to the destination. Afterwards, we apply a parameter-free strategy that transforms the initial image latent to a temporally coherent latent video by masked replacement following the 3D trajectories of both end-effector and object, achieving precise spatial control, finally, the manipulation video is synthesized with diffusion model conditioned on the latent video. 

In conclusion, our key contributions are threefold.
\begin{itemize}
    \item We propose a novel occupancy-aware 3D trajectory planner that yields collision-free, length-efficient end-effector trajectories with reasonable velocities. 
    \item We introduce a simple yet efficient trajectory-to-video synthesis scheme, which seamlessly plugs into diffusion-based models without any auxiliary modules. 
    \item Our comprehensive experiments across diverse manipulation scenarios demonstrate that \name achieves SOTA performance across multiple video quality metrics in robotic trajectory-conditioned video generation, while maintaining precise trajectory control.
\end{itemize}

%% file: sections/2_related_works.tex
\section{Related Works}
\label{sec:paper-related-works}
\subsection{Robotic Trajectory Planning}

Trajectory planning is a critical component in robotics and autonoumous driving, including vision-language navigation (VLN)~\cite{zhang2024uni, wei2025omniindoor3d, chen2025splat, chen2025streamkvstreamingvideoquestionanswering, huang2024s3g, wei2024emd} and vision-language-action (VLA) tasks~\cite{kim2024openvla, cao2025fastdrivevla}. A common paradigm of path planning in VLN involves constructing a global representation of the environment followed by subsequent path planning for mobile robots. For instance, \cite{wang2025instruction} proposes a method that samples path proposals via a random walk strategy and scores them using a multi-modal transformer. Alternatively, some studies leverage large language models (LLMs) for reasoning and planning based on human instructions and visual observations. NavGPT \cite{zhou2024navgpt} utilizes a supportive interaction and memory tracking framework, while \cite{chen-etal-2024-mapgpt} employs a global topological node map to enhance spatial reasoning for LLMs.

In contrast to VLN's global planning focus, the VLA field prioritizes real-time action prediction in a target-approaching manner, often generating short-term action chunks\cite{pmlr-v229-zitkovich23a, kim2024openvla, black2410pi0}. For instance, OpenVLA processes the current observation and outputs the next 16 (8) delta actions chunk under 15 Hz(5 Hz) control, while $\pi_{0}$\cite{black2410pi0} employs a 50-step action horizon. These predicted actions are subsequently refined into executable subtle executable trajectories by low-level controllers using numerical algorithms, like RRT\cite{lavalle1998rapidly} and CHOMP\cite{ratliff2009chomp}.

Unlike VLA models that predict conditional action distributions from observation history and robot states, robotic manipulation video generation models benefit from being provided with globally optimal paths that unlocks holistic semantic understanding of a task. To realize this, we propose planning complete 3D trajectories in occupancy space, optimized via a CHOMP-inspired method. We detail this trajectory planning approach in the path planning section. 

\subsection{Trajectory Controlled Video Generation}
Recent advances in motion-controlled diffusion video generation have demonstrated remarkable progress. TrailBlazer~\cite{ma2024trailblazer} introduces sparse keyframe bounding boxes to guide object motion.  DragAnything~\cite{wu2024draganything} employs 2D Gaussian maps and entity feature maps extracted by a pretrained model for generation guidance. MotionCtrl~\cite{wang2024motionctrl} control camera and object motion through Camera Motion Control Module (CMCM) and Object Motion Control Module (OMCM) respectively, while a recent work, DaS~\cite{gu2025diffusion}, proposes a unified 3D-tracking representation that unified camera and object motion by tracking 3D point movements. 

These methods are extended to robotic video generation where the core challenge lies in jointly modeling robot and object motion as cameras are typically fixed. This\&That~\cite{wang2024language} places one gaussian at the object's initial position on grasping frame and another at its target on placement frame of a 2D gaussian map video, controlling generation via ControlNet. RoboMaster~\cite{fu2025learning} divides the whole process into three phases based on the dominate moving agents, edits latents according to agents' movements and conditions with spatial-temporal convolution injection modules.

%% file: sections/3_method.tex
\begin{table}[!t]
    \centering
    \resizebox{1.0\linewidth}{!}{
        \begin{tabular}{c|c|c|c}
            \toprule
            \multirow{2}{*}{Method} & Keypoint  & Full Trajectory  & Affordance  \\
             & Control & Control & Control  \\
            \midrule
            This\&That & \cmark & \xmark & \xmark \\
            RoboMaster & \cmark & \cmark & \xmark \\
            \name & \cmark & \cmark & \cmark \\
            \bottomrule
        \end{tabular}
    }
    \caption{\textbf{Control capabilities of existing methods.} Ours \name\ is the most fine-grained by supporting keypoint, full-trajectory, and affordance control.}
    \label{tab:control-level-comparison}
    \vspace{-5mm}
\end{table}

\section{Method }
\label{sec:paper-method}
\subsection{Problem Formulation}
Given a single third-view observation image $I_0 \in \mathbb{R}^{3 \times H \times W}$ and corresponding instruction $t$, our goal is to generate a 3D-aware robot manipulation video. The pipeline consists of four main components:
1) We first construct a 3D occupancy map $O\in \mathbb{R}^{h \times w \times d}$. This discrete volumetric representation captures the spatial distribution of objects and the scene.
2) Upon the occupancy map, we plan an initial 3D trajectory $P^3_{init} \in \mathbb{R}^{N \times 3}$ via the $A^*$ algorithm and optimize it through gradient descent to obtain a shorter and smoother gripper trajectory $P^3_g$. The object trajectory $P^3_o$ is then acquired by relatively static assumptions. 
3) These 3D trajectories are projected to 2D mask maps of object and gripper $M^{obj}, M^{grip} \in \mathbb{R}^{N \times 3 \times H \times W}$ and guides the construction process of video latent $\mathcal{Z}$.
4) The output video $V \in \mathbb{R}^{N \times 3 \times H \times W}$ is conditionally generated with video latent $\mathcal{Z}$. Our method offers the most fine-grained control compared to baselines, as in Tab.~\ref{tab:control-level-comparison} and showcases in Fig.~\ref{fig:affordance-control}.

\subsection{Occupancy-aware path planning}
\label{sec:3d-occupancy-aware-trajectory-planning}
\subsubsection{3D Occupancy Map Reconstruction}
\label{sec:3d-occupancy-map-reconstruction}
\begin{figure*}[h]
    \centering
    \includegraphics[width=0.9\textwidth]{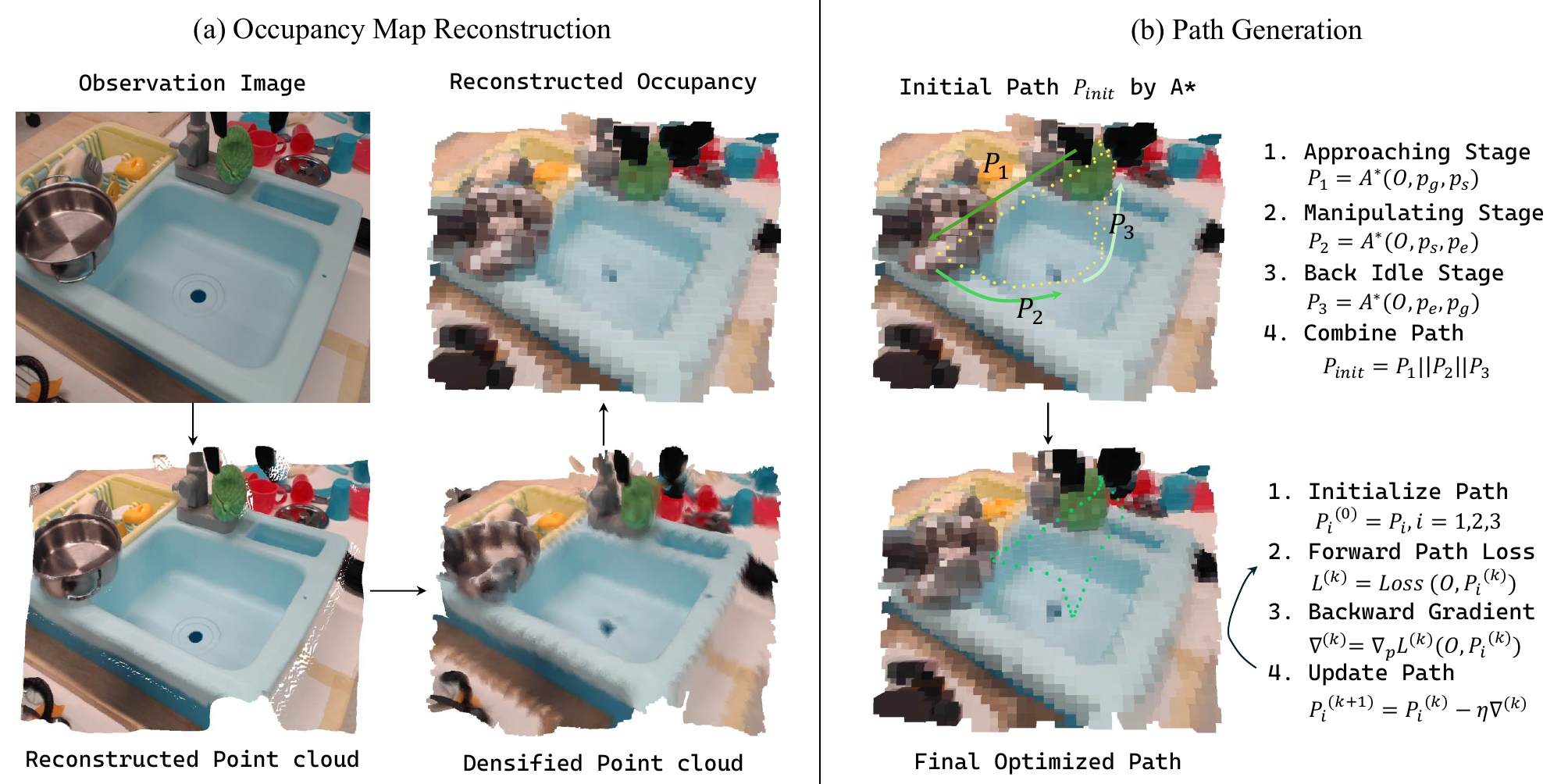}
    \caption{\textbf{The occupancy construction and trajectory pipeline.} (a) In the occupancy reconstruction process, we first estimate and densify the point cloud, then extract an occupancy out of the densified point cloud. (b) The procedure of path planning in occupancy. We first generate 3 initial sub-trajectories using $A^*$, and then each sub-trajectory is then optimized with gradient descent. We formulate the optimization process of a given path $P$ here.} 
    \label{fig:trajectory-planning}
\end{figure*}

    To enable effective robot motion planning, we first establish an accurate 3D representation of the scene. 
    We decide to use occupancy to represent the matters in the scene, as it is able to represent local structure in the scene and provides an regular scene representation that benefits  path searching. 
    We apply three steps to construct an occupancy grid given a single-view observation. Firstly, we leverage the 3D scene understanding capabilities of VGGT~\cite{wang2025vggt} to generate an initial point cloud in camera coordinates. However, this raw point cloud exhibits discontinuity in regions of occlusion. To address this, we employ a neural surface reconstruction technique~\cite{huang2023neural} that is able to recover a continuous surface from sparse points. From this reconstructed surface, we uniformly resample points to create a more complete point cloud. Finally, we discretize the 3D space by converting the processed point cloud into a $64\times64\times64$ occupancy grid, where each voxel indicates the presence or absence of matter, balancing efficiency and precision. Fig.~\ref{fig:trajectory-planning} (a) illustrates this multi-stage 3D reconstruction pipeline. 

\subsubsection{Optimal Trajectory Planning}
\label{sec:optimal-trajectory-planning}
We propose a three-stage method to find a locally optimal manipulation path in the 3D occupancy space. 
First, we generate an initial trajectory using the $A^*$ algorithm applied three times for different stages:
\begin{itemize}
    \item \textbf{Approaching stage.} The path  $P_1$  is planned from the end-effector’s 3D position to the object’s position.
    \item \textbf{Manipulating stage.} $P_2$ moves while grasping the object from its initial position to the target position.
    \item \textbf{Back-idle stage.} The last path $P_3$ returns the end effector to the start position.
\end{itemize}

While $A^*$ provides a heuristic solution, the trajectories needs to be further optimized for safety and smoothness. Therefore these sub-trajectories $P_1, P_2, P_3$ are optimized for safety and smoothness respectively.
Inspired by the widely used CHOMP algorithm, we jointly optimize the trajectory with multiple objectives to find a plausible, short, and smooth path as shown in Fig.\ref{fig:trajectory-planning}(b). Note each the $i$th point in $P^3=P^3_{init}$ as $\mathbf{p}_i$, the objectives are formulated as:
\textbf{Collision Loss.} A signed distance field (SDF) is constructed to represent the field of distance toward the static background of the scene, and a safe distance hyperparameter is adapted. 
\textbf{Path Length Loss.} This target aims to minimize the path length, ensuring the effectiveness of robot manipulation. 
\textbf{Smoothness Loss.} Smooth loss includes two parts, an acceleration loss and a curvature loss. This loss avoids sharp acceleration or direction changes of robot end effector.

\begin{figure}[!h]
    \centering
    \includegraphics[width=0.45\textwidth]{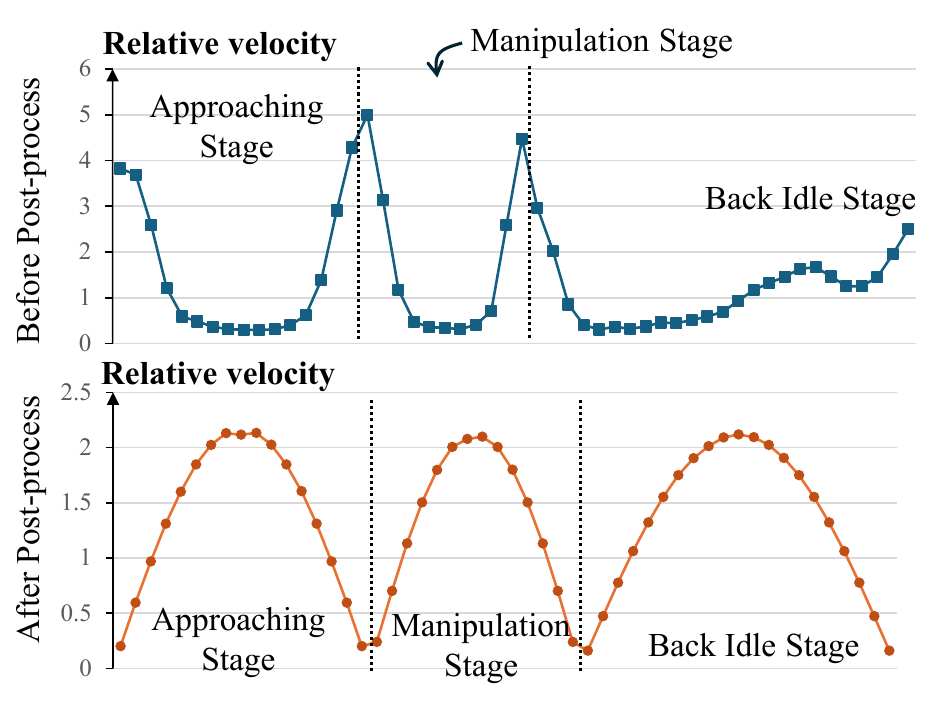}
    \caption{\textbf{Distribution of velocity before and after path-aware time reallocation of one example.}} 
    \label{fig:time-reallocation}
    \vspace{-5mm}
\end{figure}

\begin{figure*}[ht]
    \centering
    \includegraphics[width=0.9\textwidth]{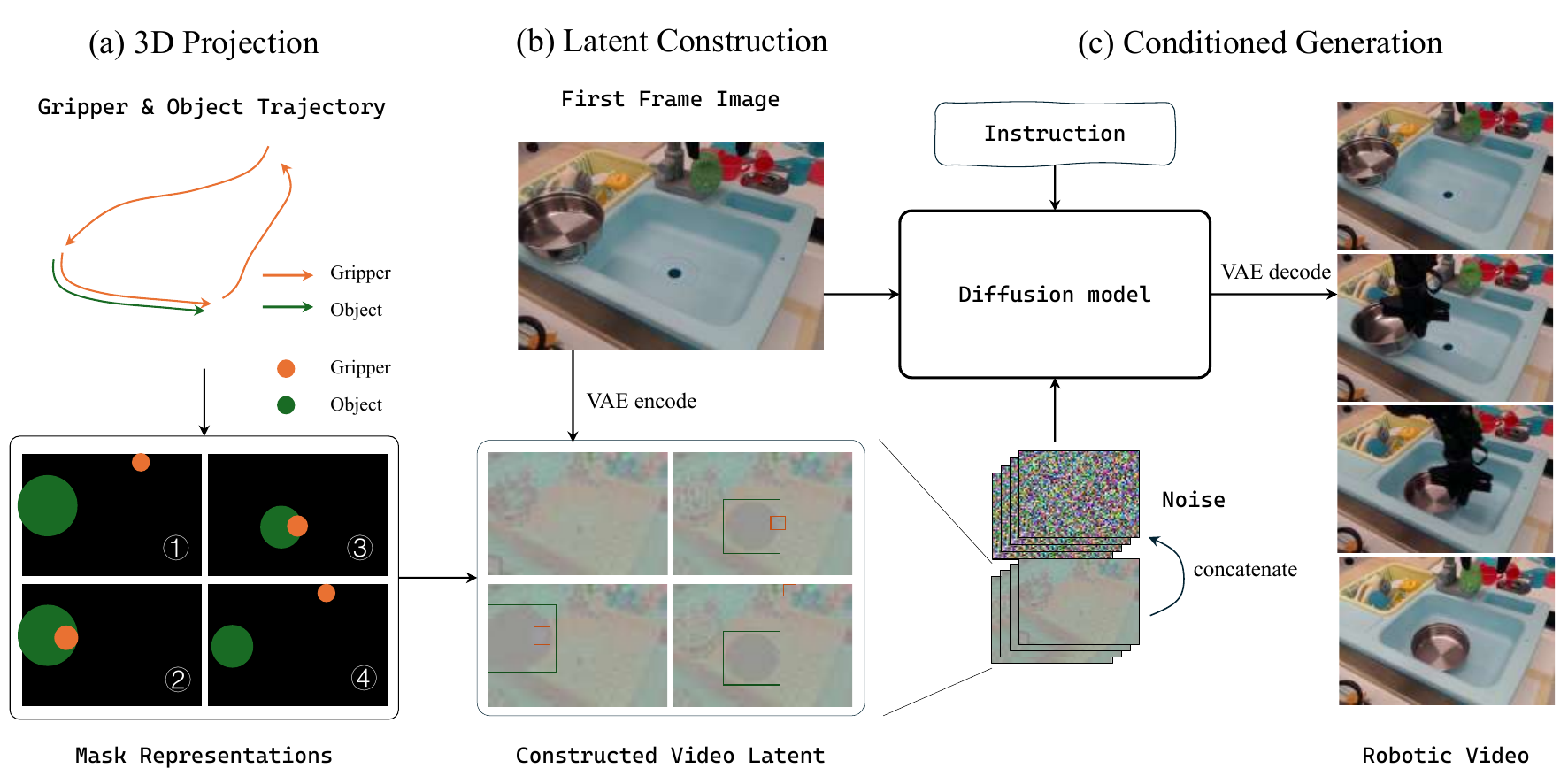}
    \caption{\textbf{The conditioned Video generation pipeline.} (a) We use a 3D-to-2D projection to create masks that represent the position and distance of object or gripper in each frame, we draw the mask of object and gripper in the same mask for clarity. (b) We apply a latent editing method using the first frame latent and the corresponding masks to create a video latent. (c) We use the constructed video latent to guide the generation of robotic videos.} 
    \label{fig:video-generation}
\end{figure*}
\vspace{-2mm}
\begin{equation}
    \mathcal{L}_{col} = \sum_{i=1}^{N}(SDF(\mathbf{p}_{i})) \label{eq-1}
\end{equation}
\vspace{-2mm}
\begin{equation}
\mathcal{L}_{len} = \sum_{i=1}^{N-1}||\mathbf{p}_i-\mathbf{p}_{i+1}||^2
\label{eq-2}
\end{equation}
\vspace{-2mm}
\begin{equation}
\mathcal{L}_{acc} = \frac{1}{2}\sum_{i=1}^{N-2}||\mathbf{p}_{i+2}-2\mathbf{p}_{i+1}+\mathbf{p}_i||^2
\label{eq-3}
\end{equation}
\vspace{-2mm}
\begin{equation}
\mathcal{L}_{cur} = \frac{1}{2}\sum_{i=1}^{N-2}\left (\frac{||\mathbf{v}_i\times \mathbf{a}_i||^2}{||\mathbf{v}_i^3||^2 + \epsilon }\right )
\label{eq-4}
\end{equation}

In the above formulation, $\mathbf{v}_i = \mathbf{p}_{i+1} - \mathbf{p}_i$ is velocity of each point,
$\mathbf{a}_i = \mathbf{p}_{i+2} - 2\mathbf{p}_{i+1} + \mathbf{p}_i$ is the acceleration term, 
$\epsilon=1e^{-6}$ is a small constant and $\times$ denotes the cross product. Note that the start and end are specially treated and are not optimized, therefore keeping their original position. We optimize the point group $P^3$ to minimize the following objective using Adam optimizer and a learning rate of 0.1 for a fixed number of iterations and finally obtain $P^3_{opt}$:
\begin{equation}
    \min_{P^3}\left( \omega_{len}\mathcal{L}_{len} + \omega_{curv}\mathcal{L}_{cur} + \omega_{acc}\mathcal{L}_{acc} + \omega_{col}\mathcal{L}_{col}\right) \label{eq-5}
\end{equation}
After the optimization process, we obtain optimized sub-trajectories $P^3_i, i=1,2,3$ that are plausible, as short and more safe in their path shape. However, due to the ignorance of the above objectives towards the mechanic nature of robot, the speed distribution of the above points follows poorly with real robots, which typically first accelerate and then decelerate within each sub-trajectory.
\subsubsection{Path-aware Time Reallocation}
To adjust the robot's speed throughout the manipulation process according to a predefined velocity profile, we propose a post-processing method that redistributes trajectory points based on path length and desired speed characteristics. First, the number of points in each sub-trajectory is reassigned proportionally to its arc length. Then, each point is repositioned along the path via interpolation between the two nearest original points, ensuring the resulting spatial distribution matches the intended velocity profile. We use sine wave as our default velocity profile for each sub-trajectory.

As shown in Fig. \ref{fig:time-reallocation}, the velocity distribution of the original trajectory exhibits inconsistencies with physics rules, whereas the recalculated trajectory after post-processing adheres more closely to physically realistic motion constraints. This realignment results in a velocity profile that is both smoother and more applicable to real-world execution.

\subsection{3D Trajectory Data Curation}
\label{sec:trajectory-data-curation}
\textbf{3D Scene Reconstruction.} We first generate temporally consistent point clouds of each frame of the third-view 2D video by processing the entire sequence with a VGGT model \cite{wang2025vggt}. This produces a unified 3D point maps of the scene along with estimated camera parameters.

\textbf{End-Effector Localization.} Accurately localizing the 3D position of the robotic end-effector is challenging due to its dynamic motion and lack of fixed reference points. To address this, we employ a fine-tuned YOLO model \cite{yaseen2408yolov8} specifically for detecting the gripper fingers. The 2D centers of the detected bounding boxes are then mapped to corresponding 3D coordinates, and the midpoint between these two positions is set as the 3D location of the gripper.
\begin{table*}[htbp]
    \centering
    \caption{\textbf{Common video metric results of our \name SVD version, DiT version, and baselines.}}
    \vspace{-2mm}
    \label{tab:video_quality_common}
    \begin{tabular}{c|c|c|c|c|c|c}
        \toprule
        \multicolumn{2}{c}{\multirow{2}{*}{Method}} & \multicolumn{3}{|c}{Video Quality} & \multicolumn{2}{c}{Trajectory Accuracy} \\
        \multicolumn{2}{c|}{}& FVD $\downarrow$  & PSNR $\uparrow$  & SSIM $\uparrow$  & $\text{TrajError}_{\text{robot}} \downarrow$ & $\text{TrajError}_{\text{obj}} \downarrow$\\
        \midrule
        \multirow{3}{*}{SVD-based}&DragAnything & 158.42 & 21.13 & 0.792 & 18.97 & 27.41 \\
        &This\&That & 148.69 & 20.93 & 0.758 & 62.07 & 37.12 \\
        &\textbf{ManipDreamer3D(SVD)} & \textbf{143.33} & \textbf{22.75} & \textbf{0.807} & \textbf{17.40} & \textbf{18.77} \\
        \midrule
        \multirow{2}{*}{DiT-based}&RoboMaster & 147.31 & 21.55 & 0.803 & 16.47 & 24.16 \\
        &\textbf{ManipDreamer3D(DiT)} & \textbf{93.98} & \textbf{23.64} & \textbf{0.847} & \textbf{15.38}& \textbf{16.59}\\
        \bottomrule
    \end{tabular}
    \vspace{-2mm}
\end{table*}

\begin{table*}[htbp]
    \centering
    \caption{\textbf{The VBench metric results of our \name SVD version, DiT version, and baselines.}}
    \vspace{-2mm}
    \label{tab:video_quality_vbench}
    \resizebox{\linewidth}{!}{
    \begin{tabular}{c|c|c|c|c|c|c|c}
        \toprule
        \multicolumn{2}{c|}{\multirow{2}{*}{Method}} & Aesthetic & Imaging & Temporal & Motion & Subject & background \\
        \multicolumn{2}{c|}{} & Quality $\uparrow$ & Quality $\uparrow$ & Flickering $\uparrow$ & Smoothness $\uparrow$ & Consistency $\uparrow$ & Consistency $\uparrow$ \\
        \midrule
        \multirow{3}{*}{SVD-based}&DragAnything & 49.53 & 67.15 & 97.83 & 98.25 & 93.01 & 95.14 \\
        &This\&That & \textbf{57.27} & \textbf{70.09} & 97.20 & 97.91 & 94.43 & 95.45 \\
        &\textbf{ManipDreamer3D(SVD)} & 52.46 & 69.24 & \textbf{97.98} & \textbf{98.47} & \textbf{95.39} & \textbf{96.57} \\
        \midrule
        \multirow{2}{*}{DiT-based}&RoboMaster & 50.32 & 67.49 & 98.27 & 98.81 & 93.55 & 95.40 \\
        &\textbf{ManipDreamer3D(DiT)} & \textbf{51.44} & \textbf{68.65} & 98.18 & 98.70 & \textbf{94.38} & \textbf{95.87} \\
        \bottomrule
    \end{tabular}
    }
    \vspace{-2mm}
\end{table*}
\textbf{Object Detection and Segmentation.} We first identify the grasping initiation moment—when the gripper is about to contact the target object. Following This\&That \cite{wang2024language}, we use the gripper’s 2D center point at this moment as 2D indication of the object’s position. We then leverage Qwen-VL \cite{bai2023qwenvlversatilevisionlanguagemodel} for visual grounding, using a prompt that combines both the estimated position and the object name (parsed from the instruction via an LLM). Finally, we apply SAM \cite{kirillov2023segment} to generate precise object masks based on the 2D position and bbox.

This multi-stage approach ensures high-fidelity 3D trajectory extraction for both the robotic arm and the manipulated object from original videos, forming a critical foundation for the training process of our video generation model.

\subsection{Trajectory-Guided Video Synthesis}
\label{sec:trajectory-guided-video-synthesis}
\subsubsection{3D-to-2D Trajectory Representation}
\label{sec:3d-to-2d-trajectory-representation}
Our framework transforms 3D manipulation trajectories into a compact 2D latent representation that effectively guides video synthesis. As shown in Fig. \ref{fig:video-generation}, the representation consists of three key components: (1) first-frame latent encoding with VAE, (2) projecting dynamic object and end-effector trajectory and construct masks, and (3) temporal latent construction. 

\textbf{Latent Editing Preparation.} We initialize our representation by extracting the latent code of the first video frame using the video diffusion model's variational autoencoder. Following RoboMaster \cite{fu2025learning}, we employ pooled latent vectors to represent key scene elements:
For the manipulated object, we interpolate its mask to match the latent spatial resolution and compute the mean-pooled latent vector within the masked region.
For the end-effector, we generate a predefined latent vectors through similar operations and add biases to represent open/closed states of the gripper.

\textbf{2D Mask Projection from 3D trajectories.} The projection process involves three computational steps:
\textit{Distance Estimation}: We establish the object's depth by assuming constant camera-object distance during grasping, using the end-effector's known 3D trajectory as a reference to estimate the change of object distance.
\textit{Geometric Abstraction}: Object and gripper are modeled as spheres, the radius is defined as the maximum edge length of the box of object, while the end-effector uses a predefined fixed small radius.
\textit{Perspective Projection}: We projection 2D circles from their 3D spheres for both object and end-effector using standard perspective projection equations, thus the scale reflects distance.

\textbf{Temporal Latent Construction.} The dynamic representation is constructed in the following way: For the first frame, we simply maintain the original latent to keep the initial spatial and semantic representation.
For the following frames, we overlay the object and gripper latent vectors within their masked regions sequentially onto the first latent. 

This representation efficiently encodes pixel-level position and distance of the manipulation trajectory while still being compatible with the diffusion model's architecture. The experimental results in the Experiments section demonstrate its effectiveness in guiding realistic video synthesis.

\subsubsection{Conditioned Video Generation}  
\label{sec:latent-editing-video-generation}  
Unlike previous trajectory conditioned methods that rely on additional ControlNet \cite{wang2024language} or injection modules \cite{fu2025learning} to bridge trajectory conditions with video features, our approach directly concatenates the constructed latent with the noisy video latent along the channel dimension, replacing the traditional repeated static first-frame condition with a dynamic video condition, leading to precise control.  

For UNet-based models such as SVD, where the VAE performs only spatial compression, the constructed latent can be directly fed into the denoising model. However, for DiT-based architectures like CogVideoX-5B \cite{yang2024cogvideox}, which uses a VAE that employs a $4\times$ temporal compression aside from spatial compression, we simply apply a two-layer $2\times$ average pooling on temporal dimension for alignment. 

This latent-editing strategy offers two key advantages: it maintains the original diffusion model framework by replacing the repeated first-frame latent with our dynamically edited version, introducing no additional parameters. Furthermore, by using the first-frame latent as the editing base, we minimize distribution shift while preserving the inherent condition mechanism of model and achieve precise control.

%% file: sections/4_experiments.tex
\section{Experiments}
\label{sec:paper-experiments}
\subsection{Experiment Setups}
\subsubsection{Model and data details.}We train our models on a combination of the bridge V1 and bridge V2 datasets. After our curation pipeline, we finally acquired 8.7k valid episodes. These examples are randomly split by 9:1 as the training set and test set. We implement our condition video generation method based on both pretrained SVD model~\cite{blattmann2023stable} and CogVideoX-5B~\cite{yang2024cogvideox}, we denote these two versions of our models as \name(SVD) and \name(DiT) respectively. 
\subsubsection{Metrics.} To evaluate the quality of generated videos, we use common video qualities including FVD~\cite{unterthiner2018towards}, SSIM~\cite{wang2004image}, and PSNR~\cite{hore2010image}, as well as VBench~\cite{huang2024vbench} metrics. Aside from that, we also evaluate our model with trajectory error, which reveals how well the model follows the trajectory. Following ~\cite{fu2025learning}, we evaluated this metric for the robot end-effector and the target object, respectively.

\begin{figure}[htbp]
    \vspace{-5mm}
    \centering
    \includegraphics[width=0.45\textwidth]{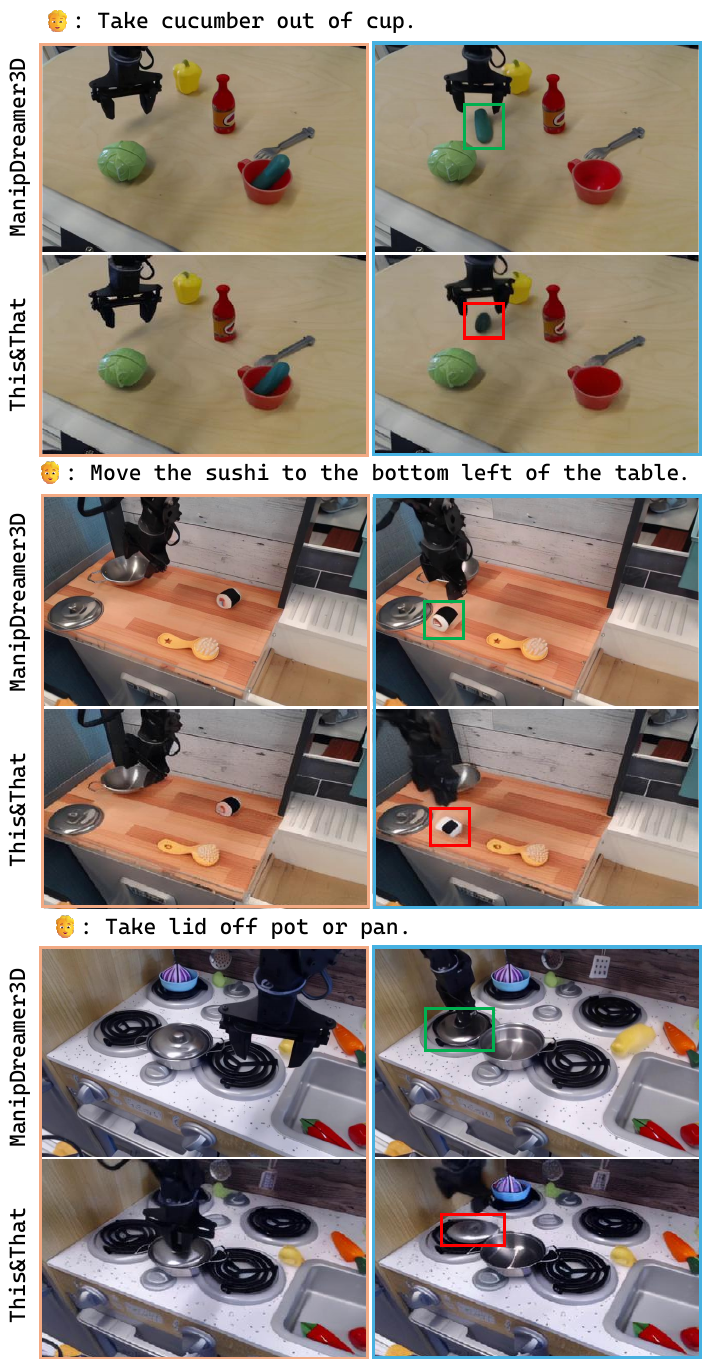}
    \caption{\textbf{Qualitative comparison between our \name and This\&That (both SVD-based)}. Our method better preserves object appearance compared to baseline This\&That, which exhibits noticeable shape distortions in manipulation results.}
    \label{fig:svd-comparison}
\end{figure}
\begin{figure}[htbp]
    \centering
    \includegraphics[width=0.47\textwidth]{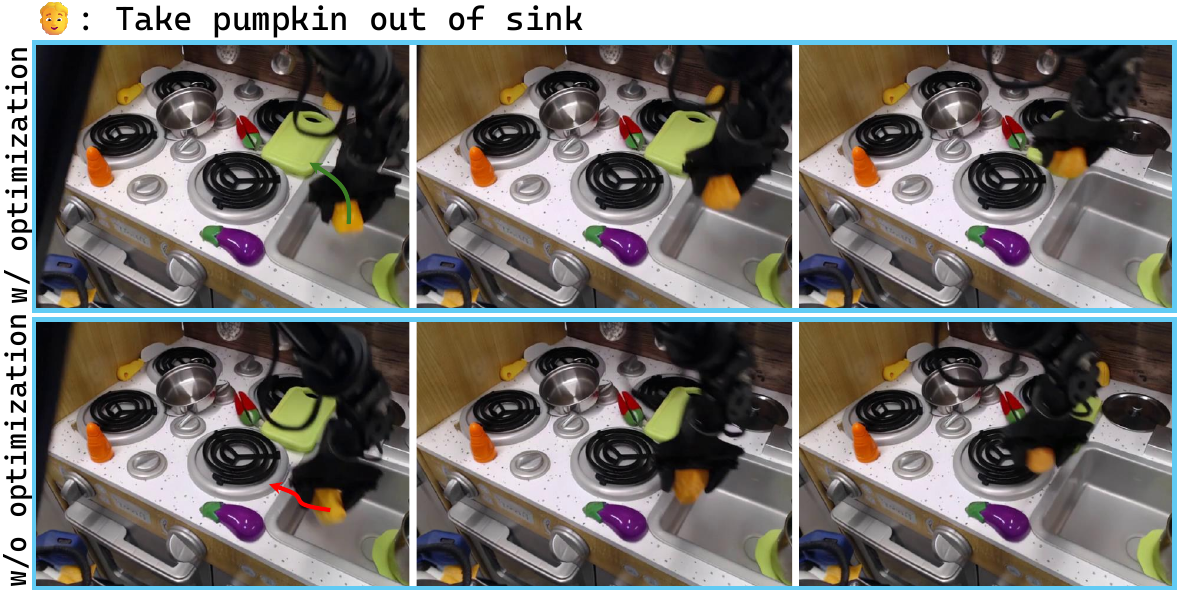}
    \vspace{-2mm}
    \caption{\textbf{Video synthesis comparison between using initial trajectories and optimized trajectories.}} 
    \vspace{-2mm}
    \label{fig:trajectory-plan-importance}
\end{figure}
\begin{figure}[ht]
    \centering
    \includegraphics[width=0.47\textwidth]{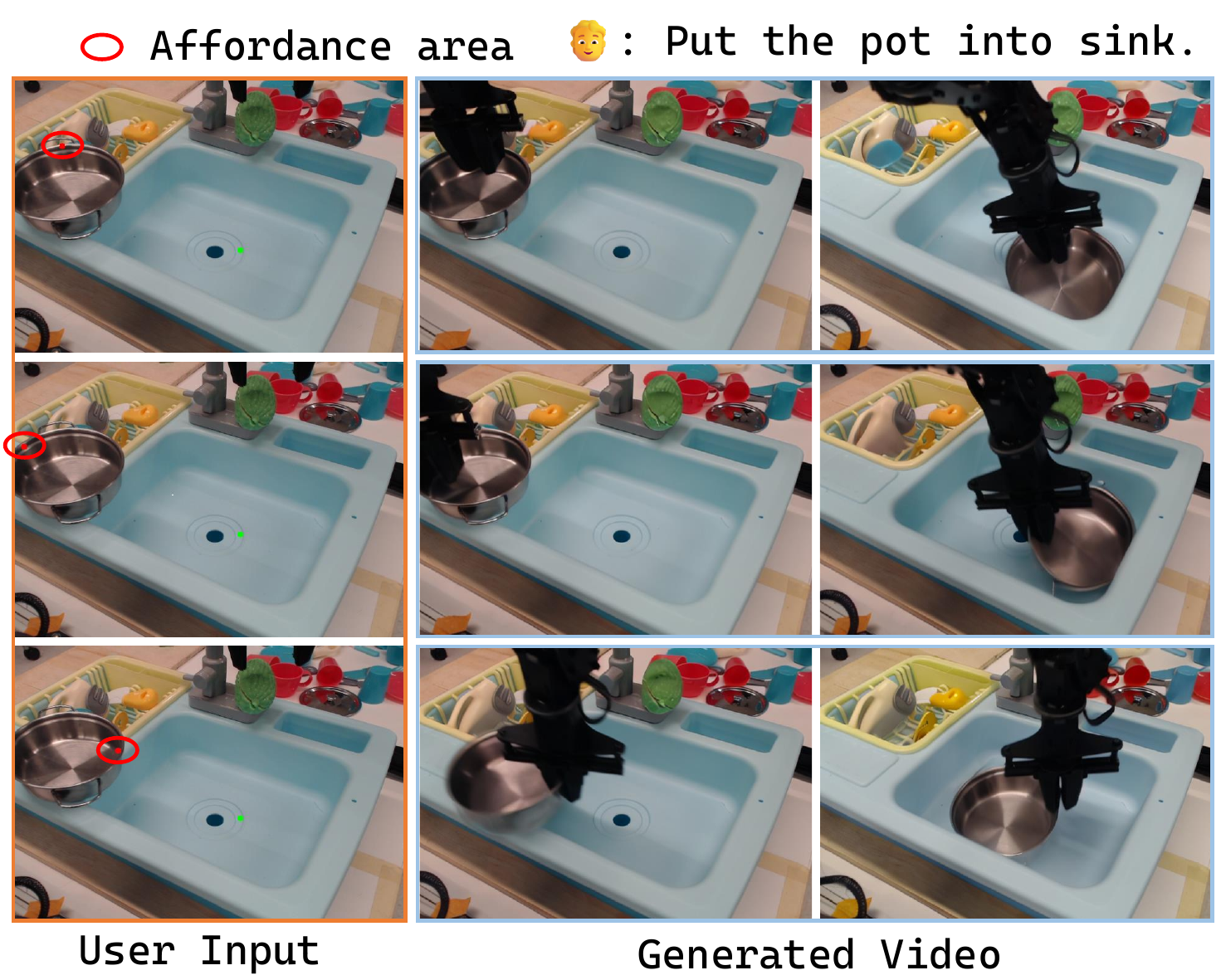}
    \vspace{-2mm}
    \caption{\textbf{Precision manipulation control demonstrations.} Our \name exhibits fine-grained control capabilities by generating manipulation videos conditioned on specific object affordances.} 
    \vspace{-2mm}
    \label{fig:affordance-control}
\end{figure}

\subsection{Video Quality Results}
The quantitative comparisons of video generation quality are presented in Tab. \ref{tab:video_quality_common} and Tab.~\ref{tab:video_quality_vbench}. Our ManipDreamer3D demonstrates superior performance across both evaluation frameworks.
In common video metrics (Tab. \ref{tab:video_quality_common}), our SVD variant achieves best-in-class scores with FVD (143.33), PSNR (22.75), and SSIM (0.807), while significantly reducing trajectory errors (17.40 for robot, 18.77 for object) compared to SVD-based competitors including DragAnything~\cite{wu2024draganything} and This\&That~\cite{wang2024language}. The DiT version further elevates performance, attaining 93.98 FVD and 0.847 SSIM with notably precise trajectory adherence (15.38 for robot and 16.59 for object). 

The trajectory accuracy advantage stems from our gripper-object collaborative representation design. While This\&That relies on ambiguous two-key-gesture control and DragAnything operates at the whole-entity movement level, lacking end-effector precision, our method explicitly models the spatial position of both gripper and object throughout the interaction. RoboMaster, which uses the same DiT backbone, implicitly models the position of the end-effector during interaction, underperforms in trajectory accuracy. 

VBench evaluation results in Tab.~\ref{tab:video_quality_vbench} reveals our method's balanced strengths: the SVD model leads in temporal stability (97.98 flickering score) and consistency metrics (95.39 subject, 96.57 background), while the DiT variant maintains competitive Imaging quality (68.65) with robust motion handling (98.70 smoothness). These results collectively validate our approach's effectiveness in both visual quality and trajectory fidelity. We also visualize some of the same examples generated by our \name(SVD) and This\&That. As shown in Fig.~\ref{fig:svd-comparison}, our model keeps the original shape of the original objects, while This\&That suffers from object deformation. This qualitative advantage aligns with our quantitative metrics, showing our superior visual quality. 

\subsection{Key Components Analysis}
\subsubsection{Precision Manipulation Control}

To evaluate the model's capability in precision manipulation control, we conduct experiments using different grasp parts (affordances) of objects to generate manipulation videos. The experiments employ our SVD-based model architecture to demonstrate fine-grained control over manipulation actions.

As illustrated in Fig.~\ref{fig:affordance-control}, we systematically vary the contact points on a pot object to test the model's responsiveness to different manipulation positions. The results show that our model successfully generates manipulation videos that accurately follow the specified conditions, demonstrating its ability to handle intricate manipulation tasks.

\subsubsection{The impact of trajectory optimization}
To evaluate the impact of trajectory optimization, we compare videos generated using initial path $P^3_{init}$ and optimized path $P^3_{opt}$ by our optimization method. As shown in Fig.~\ref{fig:trajectory-plan-importance}, with the $P^3_{opt}$ trajectory, the generated video safely avoids colliding by moving upward, while $P^3_{init}$ produces an possibly unsafe path close to the sink edge. Direct usage of videos from initial trajectory may cause unsafe behaviors in downstream VLA models, confirming the importance to optimize trajectories.

%% file: sections/5_conclusions.tex
\section{Conclusion}
We propose \name, a framework for generating robotic manipulation videos guided by 3D occupancy-aware trajectories. First, we reconstruct the scene in 3D, then plan efficient gripper and object trajectories, and finally synthesize a coherent video from the first-frame latent. This design enables comprehensive control at keypoint, full-trajectory, and affordance levels while requiring minimal manual annotation. 
Extensive experiments on diverse scenes demonstrate that \name\ improves visual quality and spatial consistency, achieving stronger trajectory adherence compared with prior motion-controlled video generators. 
\paragraph{Limitations and future work.}
Our current planning primarily targets rigid-body interactions and quasi-static grasps. Future work will explore more contact- and compliance-aware objectives, together with generative priors that better capture articulation and deformation. 